\documentclass[letterpaper,conference]{IEEEtran}
\usepackage[letterpaper,left=0.75in,right=0.75in,bottom=0.75in,top=0.75in]{geometry}
\IEEEoverridecommandlockouts
\usepackage{cite}
\usepackage{hyperref}
\usepackage{amsmath,amssymb,amsfonts}

\usepackage{algorithmic}
\usepackage{graphicx}
\usepackage{textcomp}
\usepackage{xcolor}
\usepackage{color,soul} % TODO REMOVE
\def\BibTeX{{\rm B\kern-.05em{\sc i\kern-.025em b}\kern-.08em
    T\kern-.1667em\lower.7ex\hbox{E}\kern-.125emX}}
\begin{document}

\title{\vspace*{0.25in}On the spatial attention in spatio-temporal graph convolutional networks for skeleton-based human action recognition}

\author{\IEEEauthorblockN{Negar Heidari and Alexandros Iosifidis}
\IEEEauthorblockA{\textit{Department of Electrical and Computer Engineering, Aarhus University, Denmark}\\
\{negar.heidari,ai\}@ece.au.dk}
}

\maketitle

\begin{abstract}
Graph convolutional networks (GCNs) achieved promising performance in skeleton-based human action recognition by modeling a sequence of skeletons as a spatio-temporal graph. Most of the recently proposed GCN-based methods improve the performance by learning the graph structure at each layer of the network using spatial attention applied on a predefined graph Adjacency matrix that is optimized jointly with model's parameters in an end-to-end manner. In this paper, we analyze the spatial attention used in spatio-temporal GCN layers and propose a symmetric spatial attention for better reflecting the symmetric property of the relative positions of the human body joints when executing actions. We also highlight the connection of spatio-temporal GCN layers employing additive spatial attention to bilinear layers, and we propose the spatio-temporal bilinear network (ST-BLN) which does not require the use of predefined Adjacency matrices and allows for more flexible design of the model. Experimental results show that the three models lead to effectively the same performance. Moreover, by exploiting the flexibility provided by the proposed ST-BLN, one can increase the efficiency of the model.
\end{abstract}

%\begin{IEEEkeywords}
%early exits, multi-exit architectures, curriculum learning, dynamic inference, deep learning, edge computing
%\end{IEEEkeywords}

\section{Introduction}

%Graph Convolutional Networks (GCN) attracted a great attention in recent years by generalizing the notion of convolution from grid data to graph data structures and they have been very successful in modeling non-Euclidean data structures in Computer Vision tasks such as human action recognition \citep{yan2018spatial, shi2019skeleton_directed, shi2019two, peng2020learning}. 
Considering the availability of depth cameras and successful pose estimation toolboxes such as OpenPose \cite{cao2017realtime}, each action can be represented by a sequence of body poses, each of which can be represented by a skeleton. 
Compared to other data modalities which are used for human action recognition, such as RGB videos, depth images and optical flow, human body skeleton represents the body pose and motion in a compact graph structure which is robust to context noise and invariant to body scale, view point variations, lighting conditions and background context \cite{han2017space}. 
Thus, skeleton-based human action recognition has become of great interest in recent years. Many of the recently proposed methods either use skeletons data in conjunction with other data modalities, such as RGB images \cite{franco2020multimodal}, or only utilize the skeleton data to efficiently extract high level features for human action recognition \cite{luvizon2017learning, saggese2019learning, li2017graph}.
Many deep learning methods have been proposed recently for skeleton-based human action recognition which are mainly categorized into Recurrent Neural Network (RNN)-based, Convolutional Neural Network (CNN)-based, and Graph Convolutional Network (GCN)-based methods. RNN-based methods are the earliest methods proposed in this field which mostly utilize Long Short-Term Memory (LSTM) networks \cite{greff2016lstm} to model the temporal dynamics of the sequence of skeletons \cite{du2015hierarchical,liu2016spatio,shahroudy2016ntu,song2017end,zhang2017view,li2018skeleton}. These methods represent each skeleton in a sequence as a vector which is formed by concatenated human body joints' coordinates. 
The CNN-based methods \cite{liu2017two,kim2017interpretable,ke2017new,liu2017enhanced,li2017skeleton,li2017skeletonCNN} employ the state-of-the-art CNN methods to extract the spatial and temporal features of the sequence of skeletons and they represent each skeleton in a sequence as a pseudo-image which is formed by reorganizing the body joints' coordinates into a $2$D matrix. 
Since RNN-based and CNN-based methods convert the skeleton data, which has an irregular structure, into a regular sequence or grid structure, it cannot benefit from the non-Euclidean structure of the skeleton sequences.

Recently, several GCN-based methods have been proposed for skeleton-based human action recognition and they achieved state-of-the-art performance \cite{yan2018spatial, shi2019skeleton_directed, shi2019two,peng2020learning} by utilizing the graph structure of the skeleton data. In skeleton-based human action recognition, the temporal dynamics of each action are represented by a sequence of skeletons and each human body skeleton is modeled as a graph which encodes the spatial structure of human body joints and their natural connections. 
Spatio-temporal GCN (ST-GCN) method \cite{yan2018spatial} is the first GCN-based method proposed for skeleton-based human action recognition which receives a sequence of skeletons as input and employs GCN layers to capture both spatial and temporal features in actions by exploiting the graph structure of input data. The performance of ST-GCN method has been improved by several methods which build on top of ST-GCN. These methods mostly try to adaptively learn the graph structure in each GCN layer in an end-to-end manner, based on the features of input data \cite{shi2019two, peng2020learning,shi2019skeleton_directed}.
2s-AGCN \cite{shi2019two} learns the graph structure adaptively based on the graph joints' similarity in the input data and also the existing physical connections in the body. GCN-NAS \cite{peng2020learning} is a neural architecture search method which explores the search space with different graph modules to improve the representational capacity of the GCN layers. 
Moreover, the attention mechanism has been employed in these methods in order to highlight the most important connections in body skeleton for each action class. 
AS-GCN \cite{li2019actional} extended the skeleton graphs to represent both structural links and actional links and proposed an actional-structural graph convolutional network, which has an encoder-decoder structure, to capture richer dependencies from actions. 
DPRL+GCNN \cite{tang2018deep} and TA-GCN \cite{negarTAGCN} select the most informative skeletons in a sequence to make the inference process more efficient.

In this paper, we analyze the attention mechanisms of existing spatio-temporal GCN networks used for human action recognition. Based on this analysis, we make two contributions. First, we argue that the attention mechanism in GCN layers using an additive formulation should lead to a symmetric attention mask, as the learned attentions corresponding to a pair of nodes should reflect the symmetric relationship of the corresponding human body joint positions in a human body pose. We propose a symmetric attention mechanism that is shown to perform on par with the original one for different structures of the model. Second, we propose a spatio-temporal bilinear layer which allows for more flexible design of the model for skeleton-based human action recognition. We show that models with spatio-temporal bilinear layers perform on par with spatio-temporal GCN networks without requiring the design of graph structures to encode relationships of the joints of the body skeletons.

The remainder of the paper is organized as follows. Section \ref{sec:ST-GCN} introduces the ST-GCN method which is the background of our work. Section \ref{sec:AttentionSTGCN} describes the spatial attention used in GCN-based human action recognition and Section \ref{sec:SymmetricAttentionSTGCN} analyzes its properties and proposes an alternative symmetric spatial attention. Section \ref{sec:ST-BLN} describes the spatio-temporal bilinear network, which is motivated by the analysis of the spatial attention in GCN-based human action recognition methods. The experimental results are provided in section \ref{sec:experiments} and the conclusions are drawn in section \ref{sec:conclusion}.

\section{Spatio-Temporal GCN (ST-GCN)}\label{sec:ST-GCN}
\begin{figure}
    \centering
    \includegraphics[height=0.45\linewidth]{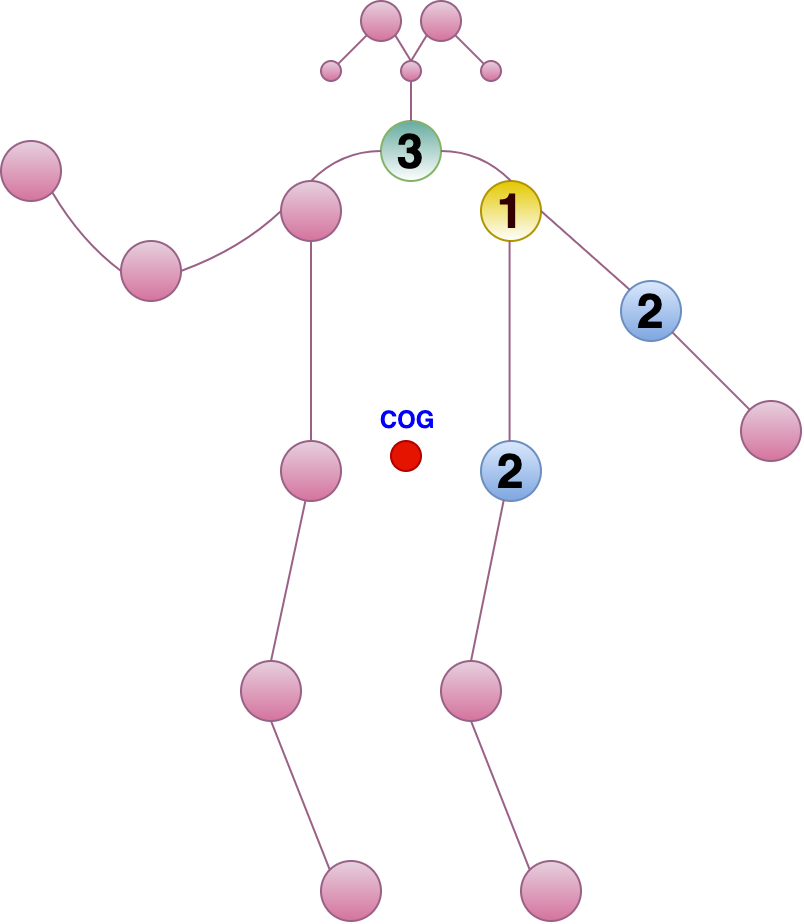}
    \includegraphics[height=0.48\linewidth]{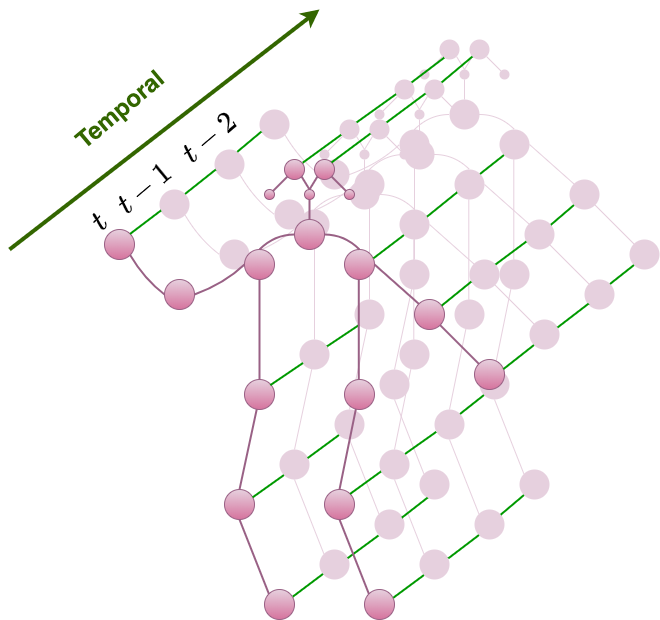}
    \caption{Illustration of an example Spatio-temporal graph (right), and the neighboring subsets in spatial partitioning process in different colors (left).}
    \label{fig:ST-Graph-Partitioning}
\end{figure}

In this section, the ST-GCN method \cite{yan2018spatial} is introduced as the baseline of many recent GCN-based methods for skeleton-based human action recognition. 
This method receives a sequence of skeletons $\mathbf{X} \in \mathbb{R}^{C^{in} \times T \times V}$ encoding the human body poses comprising an action as input, where $C^{in}$ is the number of input channels, $T$ is the number of skeletons in the sequence, and $V$ is the number of joints in each skeleton. It then models the sequence of skeletons as a spatio-temporal graph and applies multiple GCN layers with spatial and temporal convolutions to extract high level features for classifying the input skeleton sequence to a set of pre-defined action classes. 
A spatio-temporal graph on a sequence of skeletons is denoted as $\mathcal{G} = (\mathcal{V} ,\mathcal{E})$, where $\mathcal{V}$ is the set of the human body joints and $\mathcal{E}$ denotes both spatial and temporal edges connecting the body joints within each skeleton and between skeletons. The spatial edges are the natural physical connections between human body joints and the temporal edges connect each body joint of a skeleton to its corresponding joint of previous and subsequent skeletons. Therefore, the spatial graph is constructed based on the human body structure in which each node can have different numbers of neighbors, while in the temporal graph each node has two fixed neighbors, i.e. the nodes corresponding to the same joint in the previous and next time step. Fig. \ref{fig:ST-Graph-Partitioning} (right) shows the spatio-temporal graph on a sequence of skeletons. 

Since the body motions can be grouped to concentric and eccentric, ST-GCN employs a spatial partitioning process to divide the neighboring set of each node into three subsets based on their spatial arrangement in a single skeleton. This partitioning process fixes the node degrees in the spatial graph and defines the partitions as: 1) the root node itself, 2) the root node's neighbors which are closer to the skeleton's center of gravity than the root node and 3) the remaining root node's neighbors that are farther from the skeleton's center of gravity.
The skeleton center of gravity (COG) is denoted as the average of all skeleton joints' coordinates. In Fig. \ref{fig:ST-Graph-Partitioning} (left) a spatially partitioned graph is illustrated in which the nodes in different neighboring subsets (partitions) are shown with different colors and the center of gravity is shown as a red dot.  

According to the partitioning process, the graph structure in each skeleton is captured by three binary Adjacency matrices, which are hereafter indexed by $p$, each of which encodes the structure of a neighboring node subset. 
The Adjacency matrix of the first partition (corresponding to the root nodes) indicates the nodes' self-connections and is set to $\mathbf{A}_1 = \mathbf{I}_V$. Each adjacency matrix is normalized as $\hat{\mathbf{A}}_{p} = \mathbf{D}^{-\frac{1}{2}}_p\mathbf{A}_p\mathbf{D}^{-\frac{1}{2}}_p$, where $\mathbf{D}_{({ii})_p} = \sum_{j}^{V} \mathbf{A}_{({ij})_p} + \varepsilon$ denotes the degree matrix and $\varepsilon = 0.001$ is used to avoid division with zero that can be caused by empty rows in $\mathbf{D}_{({ii})_p}$. The element $\hat{\mathbf{A}}_{p,ij}$ indicates whether the vertex $j$ is in the $p^{th}$ neighboring subset of vertex $i$.

Each GCN layer updates the human body features through both spatial and temporal domains by applying spatial and temporal convolutions on the output of the previous layer. The spatial convolution is defined based on the layer-wise propagation rule of GCNs \cite{kipf2016semi} and the $l^{th}$ layer spatial convolution is given by:
\begin{equation}
    \mathbf{H}_s^{(l)} = ReLU\left(\sum_{p} \left(\mathbf{\hat{A}}_p\otimes \mathbf{M}_p^{(l)} \right)\mathbf{H}^{(l-1)}\mathbf{W}_p^{(l)} \right),
    \label{eq:2dConv_s}
\end{equation}
where $\otimes$ is the element-wise product of two matrices and $\mathbf{H}^{(l-1)} \in \mathbb{R}^{C^{(l-1)} \times T^{(l-1)} \times V}$ denotes the output of the ${(l-1)}^{th}$ layer which is introduced to the $l^{th}$ layer as input. The input of the first layer is defined as $\mathbf{H}^{(0)} = \mathbf{X}$. 
Since there are three spatial graphs for each sample, the GCN's propagation rule is applied on each graph with a different weight matrix $\mathbf{W}^{(l)}_p \in \mathbb{R}^{C^{(l)} \times C^{(l-1)}}$, where  $C^{(l-1)}$ and $C^{(l)}$ denote the number of channels in layers $l-1$ and $l$, respectively. $\mathbf{M}^{(l)}_p \in \mathbb{R}^{V \times V}$ is a learnable attention matrix which highlights the most important connections in a skeleton for each action. It is initialized as an all-one matrix and is element-wise multiplied to its corresponding adjacency matrix of each graph partition.  
The spatial convolution is in practice a 2D convolution operation which performs $C^{(l)}$ convolutions with filters of size $C^{(l-1)} \times 1 \times 1$ on the input tensor and the resulting output is a tensor of size $C^{(l)} \times T^{(l-1)} \times V$ which is multiplied with the masked attention-based adjacency matrix $\mathbf{\hat{A}}_p\otimes \mathbf{M}_p^{(l)}$ on the last dimension $V$. 

To benefit from the motions taking place in each action for label prediction, 
the output of the spatial convolution $\mathbf{H}_s^{(l)}$ is introduced to the temporal convolution block which is also a $2$D convolution that applies $C^{(l)}$ convolutions with filters of size $C^{(l)} \times k \times 1$ to capture the temporal dynamics in the sequence of skeletons by propagating the information through the time domain while keeping the number of channels unchanged. 
Based on the filter size $k$, the temporal convolution propagates the features through $k$ consecutive skeletons and reduces the number of skeletons or keeps it unchanged depending on the stride and padding size. Accordingly, the output of temporal convolution is of size $C^{(l)} \times T^{(l)} \times V$ in which $T^{(l)}$ denotes the temporal dimension of sequence in layer $l$. 
In ST-GCN method, the kernel size for temporal convolution in each layer is set to $k = 9$.
The extracted spatio-temporal features of the last ST-GCN layer are introduced to a fully connected classification layer which is equipped by a global average pooling block and a SoftMax activation function. The entire model is trained in an end-to-end manner using Backpropagation. 

\section{The role of spatial attention in GCN-based human action recognition}\label{sec:AttentionSTGCN}
One of the main drawbacks of ST-GCN method which is addressed by more recently proposed methods, such as 2s-AGCN \cite{shi2019two}, is that it uses a fixed graph structure which is heuristically predefined and represents the natural physical connections between the body joints in a skeleton. The attention mechanism in ST-GCN, by using an element-wise multiplication between the Adjacency matrix and the learnable attention mask, can only highlight or diminish existing connections between the body joints which is not guaranteed to be optimal for action classification task. For some actions such as “touching head”, the connection between the hand and head is important for recognizing the action, while this connection does not exist naturally in the body and, thus, is not considered in the predefined graph structure. 
Therefore, 2s-AGCN defines the spatial convolution as follows: 
\begin{equation}
    \mathbf{H}_{s}^{(l)} = ReLU\left(\sum_{p} \left(\mathbf{\hat{A}}_p + \mathbf{M}_{p}^{(l)} \right)\mathbf{H}^{(l-1)}\mathbf{W}_{p}^{(l)} \right),
    \label{eq:A2dConv_s}
\end{equation}
where $\mathbf{M}_{p}^{(l)}$ is a learnable attention matrix added to the graph in order to both highlight/diminish existing connections between the skeleton joints and also add potentially important connections between the disconnected ones. This matrix is initialized by zeros and it is learned in an end-to-end manner along with the other model's parameters. In other words, the spatial graph is only initialized by the skeleton's structure, and it is updated adaptively by an attention matrix whose parameters are learned in the training process. 
Comparing the attention-based Adjacency matrices of ST-GCN and 2s-AGCN, $\mathbf{\hat{A}}_p\otimes \mathbf{M}_p^{(l)}$ and $\mathbf{\hat{A}}_p + \mathbf{M}_p^{(l)}$ respectively, it can be seen that the former is guaranteed to have a structure encoding the natural connections between the human body joints, while the latter corresponds to a matrix encoding a fully-connected graph structure. 

\section{Symmetric spatial attention for GCN-based human action recognition}\label{sec:SymmetricAttentionSTGCN}
We focus on the properties of the attention-based Adjacency matrix $\mathbf{\hat{A}}_p + \mathbf{M}_{p}^{(l)}$, as it has been shown to improve action classification performance compared to the one given by $\mathbf{\hat{A}}_p\otimes \mathbf{M}_p^{(l)}$. It can be shown that our analysis also holds for the latter case. Since the matrix $\mathbf{M}_{p}^{(l)}$ is optimized in an end-to-end manner jointly with the parameters of the entire network without imposing any constraints, as detailed in Eq. (\ref{eq:A2dConv_s}), its final form will be an asymmetric matrix containing both positive and negative values. This form of the attention mask can be qualitatively explained by considering that the attention put by a graph node $v_i$ to another graph node $v_j$ can be freely optimized and can differ from the attention put by $v_j$ to $v_i$. However, considering that the nodes of the graph correspond to human body skeleton joints and during action execution their relative positions are symmetric, we would expect that their pair-wise attentions should be the same. In order to enforce this property in the optimization process of the attention-based Adjacency matrix we define the attention mask to be a symmetric matrix, i.e. $\mathbf{M}^{(l)}_p = \mathbf{L}^{(l)}_p \mathbf{L}^{T^{(l)}}_p$. Accordingly, we define the spatial convolution as follows:
\begin{equation}
    \mathbf{H}_{s}^{(l)} = ReLU\left(\sum_{p} \left(\mathbf{\hat{A}}_p + \mathbf{L}^{(l)}_p \mathbf{L}^{{(l)}^{T}}_p \right)\mathbf{H}^{(l-1)}\mathbf{W}_{p}^{(l)} \right).
    \label{eq:psd_convs}
\end{equation}
While one can select $\mathbf{M}^{(l)}_p$ to be of low rank by selecting $\mathbf{L}^{(l)}_p \in \mathbb{R}^{V \times q}$ with $q < V$, we do not set any further restrictions to $\mathbf{M}^{(l)}_p$, except of being symmetric. Once the training process ends, the matrix $\mathbf{M}^{(l)}_p$ is calculated and used for inference, thus, the space and time complexities remain the same as in the case of using Eq. (\ref{eq:A2dConv_s}). Here we should note that, since the normalized Adjacency matrices $\mathbf{\hat{A}}_p$ are designed to be asymmetric, the final attention-based Adjacency matrices will be asymmetric too. 

\begin{figure}[!t]
    \centering
    \includegraphics[width=1\linewidth]{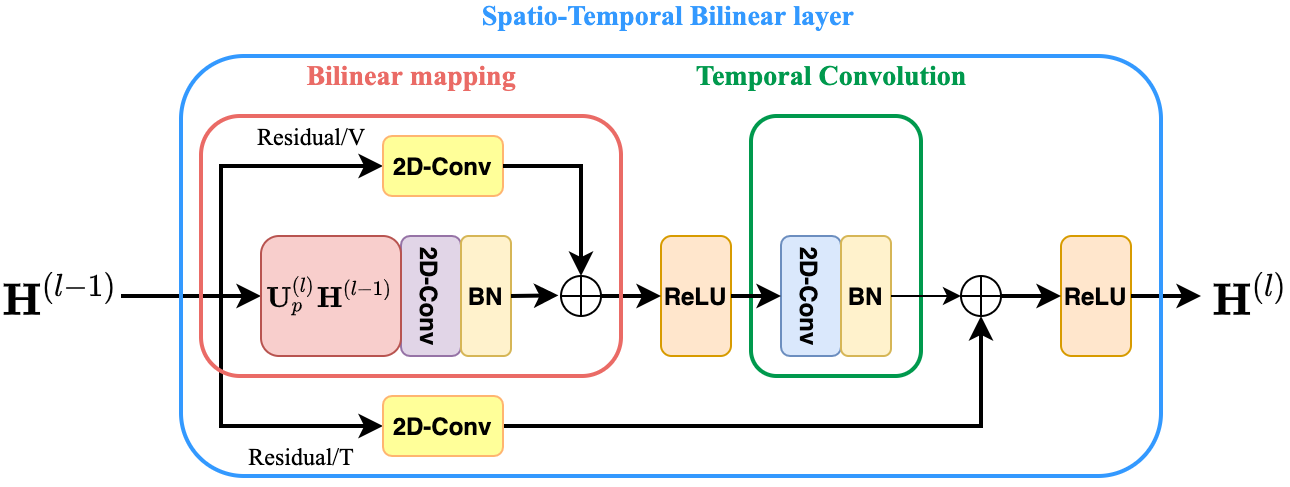}
    \caption{Illustration of spatio-temporal bilinear layer $l$ receiving as input $\mathbf{H}^{(l-1)}$ of size $C^{(l-1)} \times T^{(l-1)} \times V^{(l-1)}$ providing an output $\mathbf{H}^{(l)}$ of size $C^{(l)} \times T^{(l)} \times V^{(l)}$. 2D-Conv blocks correspond to standard 2D convolutions. The 2D-Conv block in bilinear mapping transforms the data by applying $C^{(l)}$ convolutions with filters of size  $C^{(l-1)} \times 1 \times 1 $ to change the number of channels. The 2D-Conv block in temporal convolution keeps the number of channels unchanged and applies $C^{(l)}$ filters of size $C^{(l)} \times k \times 1 $ in order to aggregate the features in temporal domain. The residual connections also apply 2D convolutions on the layer's input to have the same dimension as the layer's output. Residual/V indicates the residual connection which adds the input to the output of bilinear mapping, and Residual/T denotes the residual connection which adds the layer's input to the output of temporal convolution. 
    BN and ReLU represent the batch-normalization and ReLU activation function, respectively.}
    \label{fig:ST-BLN1}
\end{figure}

\section{Spatio-Temporal Bilinear Network}\label{sec:ST-BLN}
Considering the optimization of the attention-based Adjacency matrices based on the definition of the spatial convolution of ST-GCN in Eqs. (\ref{eq:A2dConv_s}) and (\ref{eq:psd_convs}), it can be seen that the addition of the normalized Adjacency matrix $\mathbf{\hat{A}}_p$ to the attention mask serves as an initialization strategy, while the final form of the attention-based Adjacency matrix depends primarily on the learned values of the mask. This means that, after the first few training epochs, the GCN blocks of the ST-GCN using an additive attention mask do not follow the graph structure anymore, but they are rather equivalent to bilinear layers \cite{gao2016compact,tran2018temporal,kim2018bilinear}. That is, Eqs. (\ref{eq:A2dConv_s}) and (\ref{eq:psd_convs}) can take the form:
\begin{equation}
    \mathbf{H}_{s}^{(l)} = ReLU\left(\sum_{p} \mathbf{U}^{(l)}_p\mathbf{H}^{(l-1)}\mathbf{W}_{p}^{(l)} \right),
    \label{eq:bl_Convs}
\end{equation}
where $\mathbf{U}^{(l)}_p \in \mathbb{R}^{V^{(l)} \times V^{(l-1)}}$ is a learnable matrix which is optimized in an end-to-end manner jointly with the parameters of the entire network. A specific setting of the matrix $\mathbf{U}^{(l)}_p$ for which $V^{(l-1)} = V^{(l)} = V$ corresponds to the attention-based Adjacency matrix in Eqs. (\ref{eq:A2dConv_s}) and (\ref{eq:psd_convs}) receiving as input the representations for the $V$ human body joints at layer $l$ and transforming them by applying a data transformation using $\mathbf{W}_{p}^{(l)}$ and combining information of neighboring nodes according to the connectivity in $\mathbf{U}^{(l)}_p$. However, by using the bilinear layer definition in Eq. (\ref{eq:bl_Convs}) one can freely decide on the dimensions of the transformation to be applied using $\mathbf{U}^{(l)}_p$. That is, considering that in the first layer of the Spatio-Temporal Bilinear network the matrix $\mathbf{U}^{(1)}_p \in \mathbb{R}^{V^{(1)} \times V}$ performs a transformation in the mode of the input $\mathbf{H}^{(0)}$ corresponding to the human body skeleton joints, selection of $V^{(1)} < V$ will lead to aggregation of joints' information to create a new set of of $V^{(1)}$ nodes. On the other hand, selection of $V^{(1)} > V$ will lead to the creation of new nodes to be processed by the second layer. A similar explanation can be given to the selection of the values $V^{(l)}$ for layer $l$. A special case in this setting is the one where $V^{(l)} = 1$, leading to the creation of one node encoding information of the entire input skeleton data.

Similar to ST-GCN, a spatio-temporal bilinear layer is equipped with a batch normalization, residual connections and ReLU activation function. The structure of a bilinear layer is shown in Fig. \ref{fig:ST-BLN1}. The layer has two residual connections to stabilize the model by summing up the layer's input with the layer's output. One of them adds the input of the layer to the output of bilinear mapping and the second one adds the layer's input to the output of temporal convolution. We can distinguish two cases for the residual connections. When $V^{(l-1)} = V^{(l)}$, the number of dimensions of the input tensor $\mathbf{H}^{(l-1)}$ to layer $l$ changes by applying the transformation in one of its modes using $\mathbf{W}_{p}^{(l)}$, and the residual connections need to perform a 2D convolution with $C^{(l)}$ filters of size $C^{(l-1)} \times 1 \times 1$ on the layer's input to have the same channel dimension as the layer's output $\mathbf{H}^{(l)}$. When $V^{(l-1)} \neq V^{(l)}$, the number of dimensions of the input tensor $\mathbf{H}^{(l-1)}$ to layer $l$ changes by applying the transformations in both of its modes using $\mathbf{W}_{p}^{(l)}$ and $\mathbf{U}_{p}^{(l)}$, and the residual connections need to perform a 2D convolution with $C^{(l)}\cdot V^{(l)}$ filters of size $C^{(l-1)} \times 1 \times V^{(l-1)}$ on the layer's input. The resulting tensor is then reshaped to a tensor of $C^{(l)} \times T^{(l)} \times V^{(l)}$ to have the same dimension as the layer's output $\mathbf{H}^{(l)}$. 

\begin{figure}[!t]
    \centering
    \includegraphics[width=1\linewidth]{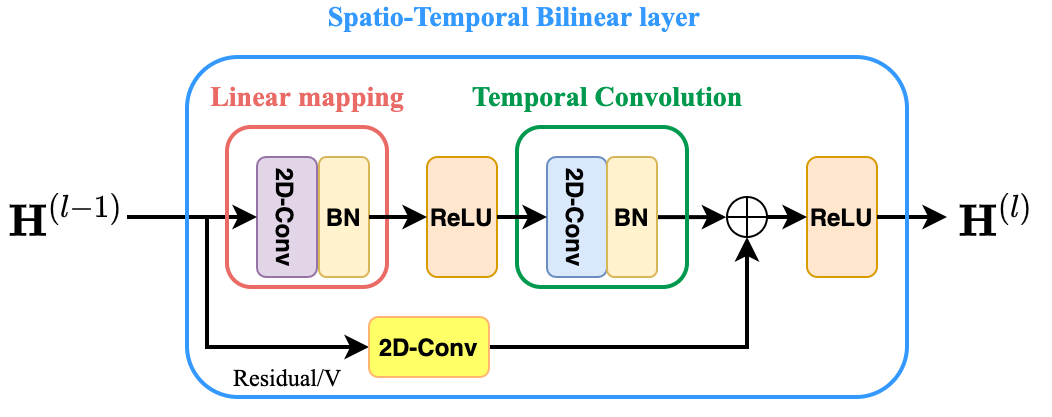}
    \caption{Illustration of spatio-temporal bilinear layer $l$ when $V^{(l-1)} = V^{(l)} = 1$.} 
    \label{fig:ST-BLN2}
\end{figure}

If at layer $l-1$ a value of $V^{(l-1)} = 1$ is used, the input $\mathbf{H}^{(l-1)}$ to layer $l$ becomes of size $C^{(l-1)} \times T^{(l-1)} \times 1$. In that case there is no need of using $\mathbf{U}^{(l)}_p \in \mathbb{R}$ as the scaling factor that would be learned by using it can be absorbed in $\mathbf{W}_{p}^{(l)}$. Thus, the bilinear mapping can be replaced with an equivalent linear mapping. Moreover, in this case there is no need of using a residual connection for the linear mapping block. The architecture of such a spatio-temporal bilinear layer is shown in Fig. \ref{fig:ST-BLN2}. 

Similar to the ST-GCN method, the output of the last spatio-temporal bilinear layer is passed to a global average pooling layer to leading to 256-dimensional features for each sequence and the resulting vector is introduced to a fully connected classification layer which is equipped with a SoftMax activation function. The entire model is trained in an end-to-end manner using Backpropagation to optimize the objective function defined on the output using the classification loss. 

\section{Experiments}\label{sec:experiments}
%\subsection{Dataset}\label{subsec:dataset}
We conducted two sets of experiments on NTU-RGB+D dataset \cite{shahroudy2016ntu} which is the largest indoor-captured action recognition dataset and it is widely used for evaluating the skeleton-based human action recognition methods. 
It consists of $56,880$ action video clips from $60$ different human action classes and each clip is captured by $3$ cameras from $3$ different views. 
Each action clip is used to extract a sequence of $300$ skeletons, each of which is represented by the $3$D coordinates of $25$ body joints. This leads to each action video clip being represented as a $3 \times 300 \times 25$ tensor.
Two benchmarks are defined in \cite{shahroudy2016ntu}, i.e. cross view (CV) and cross-subject (CS), and a train-test split for each benchmark is also provided. We use the same experimental protocols and data splits as in \cite{shahroudy2016ntu}. The CV benchmark contains $37,920$ training samples captured by cameras two and three and $18,960$ test samples captured by the first camera.
In CS benchmark, the training set contains $40,320$ samples and test set contains $16,560$ samples while the actors in training and test set are different.

The experiments are conducted on PyTorch deep learning framework \cite{paszke2017automatic} with 4 GRX 1080-ti GPUs. We used the same experimental settings and network architectures as in 2s-AGCN \cite{shi2019two}. The model contains 10 GCN layers plus one fully connected layer for classification. The first 4 GCN layers have 64 output channels which are followed by the next 3 GCN layers with 128 output channels. The number of output channels in the the last 3 GCN layers is 256 the extracted feature tensor introduced to a global average pooling layer which provides a 256 diemnsional feature vector for classification. The resulting feature vector is classified into 60 different classes by the fully connected layer. 
The model is trained for $50$ epochs with SGD optimizer and cross entropy loss function. The mini-batch size is set to $64$ and the learning rate is initialized to $0.1$ and it is divided by $10$ at epochs $30$ and $40$. 

\begin{table}
\caption{Comparison of classification accuracy of spatio-temporal networks equipped with asymmetric and symmetric spatial attentions, and the bilinear mappings on the test set of CV benchmark of NTU-RGB+D dataset. The models are trained using joints data.}\label{tab:ACC_3variants}
\begin{center}
    \begin{tabular}{l|c|c}
    \hline
    Attention/Mapping  & CS(\%) & CV(\%) \\
    \hline\hline
    $\mathbf{\hat{A}}_p + \mathbf{M}_{p}^{(l)}$ & 85.29 & 93.78  \\ 
	$\mathbf{\hat{A}}_p + \mathbf{L}^{(l)}_p \mathbf{L}^{{(l)}^{T}}_p$ & 87.14 & 93.86  \\
	$\mathbf{U}^{(l)}_p$ & 85.71 & 93.80  \\ 
    \hline
    \end{tabular}
\end{center}

\end{table}

\begin{table}
\caption{Comparison of classification accuracy of spatio-temporal networks equipped with asymmetric and symmetric spatial attentions, and the bilinear mappings on the test set of CV benchmark of NTU-RGB+D dataset for different number of layers. The models use the same layer sizes as those in \cite{shi2019two} and are trained using joints data.}\label{tab:ACC_10layers}
\begin{center}
    \begin{tabular}{c|c|c|c}
    \hline
    \#Layers  & $\mathbf{\hat{A}}_p + \mathbf{M}_{p}^{(l)}$ & $\mathbf{\hat{A}}_p + \mathbf{L}^{(l)}_p \mathbf{L}^{{(l)}^{T}}_p$ & $\mathbf{U}^{(l)}_p$\\
    \hline\hline
    1 & 77.09 & 77.16 & 78.39 \\ 
    2 & 87.53 & 87.00 & 88.14 \\ 
    3 & 89.90 & 89.93 & 89.13 \\ 
    4 & 90.58 & 90.80 & 90.51 \\ 
    5 & 91.96 & 92.22 & 92.15 \\ 
    6 & 92.89 & 93.15 & 92.40 \\ 
    7 & 93.37 & 93.37 & 93.51 \\ 
    8 & 93.51 & 93.95 & 93.74 \\ 
    9 & 93.73 & 93.76 & 93.52 \\ 
    10 & 93.78 & 93.86 & 93.80 \\ 
    \hline
    \end{tabular}
\end{center}
\end{table}

In the first set of experiments, we compared the performance of spatio-temporal networks using GCN layers equipped with attentions in Eqs. (\ref{eq:A2dConv_s}) and (\ref{eq:psd_convs}) and using bilinear layer as defined in Eq. (\ref{eq:bl_Convs}). The performance of the three models is reported in Table \ref{tab:ACC_3variants}. As can be seen, the three networks achieved very similar performance. This indicates that there is no difference in using the predefined graph structure encoded by the matrices $\mathbf{\hat{A}}_p$ in Eqs. (\ref{eq:A2dConv_s}) and (\ref{eq:psd_convs}) for initializing the mapping in the bilinear layer (\ref{eq:bl_Convs}). To further analyze the effect of the models' size and the differences in performance achieved by the three types of spatio-temporal networks, we conducted multiple experiments using different number of layers and the results are reported in Table \ref{tab:ACC_10layers}. As can be seen, the performances of the three types of networks for all network sizes are effectively the same. This confirms our observations related to the role of the predefined graph structure used in the spatial attention of spatio-temporal networks for skeleton-based action recognition. We can conclude that the combining information related to the nodes of the human body skeleton graph can be randomly initialized and optimized jointly with the other network parameters, and that the use of predefined skeleton graph structures is not necessary for recognizing the human actions using skeleton data.
\begin{figure}[!t]
    \centering
    \includegraphics[width=\linewidth]{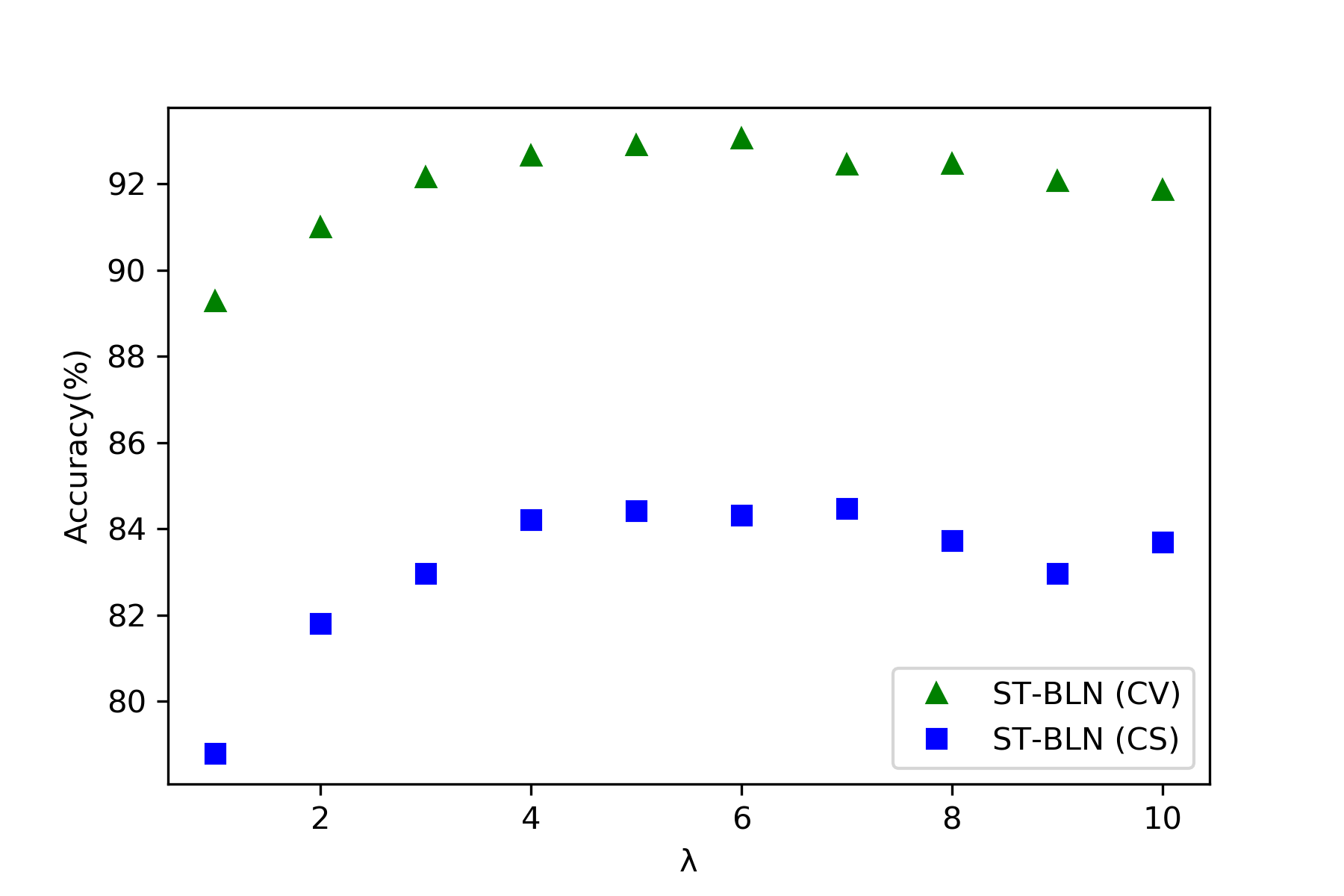}
    \caption{Performance of ST-BLN model in terms of classification accuracy, when joints' features are aggregated in different layers $\lambda$ of the network.} 
    \label{fig:ACC_BLN}
\end{figure}

For completeness, we also provide in Table \ref{table:NTU-ACC} the performance of state-of-the-art methods in skeleton-based human action recognition on the two benchmarks of the NTU-RGB+D dataset. Since most of the state-of-the-art methods, like 2s-AGCN, GCN-NAS, AS-GCN and 2s-TA-GCN, train their model with two or more data streams, we also report the performance achieved by the spatio-temporal bilinear network using two streams. That is, the 2s-ST-BLN model is formed by two streams, the first of which receives as input the joints data and the second one receives as input the bone data. The two streams process their inputs and the predicted SoftMax scores of the two streams are fused to obtain the final classification outcome.
The results show that ST-BLN and 2s-ST-BLN perform on par with other state-of-the-art methods while they do not require a predefined graph structure to encode the relationships between the joints (and bones) of the human body skeleton. Similar to the GCN-based methods, the proposed method also outperforms the RNN-based and CNN-based methods with a large margin.
\begin{table}
\caption{Classification accuracy comparison of spatio-temporal bilinear model with state-of-the-art GCN-based methods on the test set of NTU-RGB+D dataset.}\label{table:NTU-ACC}
\begin{center}
\resizebox{\linewidth}{!}{
    \begin{tabular}{l|c|c|c}
        \hline
        Method  & CS(\%) & CV(\%) & \#Streams \\ %
        \hline
        HBRNN \cite{du2015hierarchical}  & 59.1 & 64.0 & 5  \\ %\cite{du2015hierarchical}
		Deep LSTM \cite{shahroudy2016ntu}  & 60.7 & 67.3 & 1  \\ %\cite{shahroudy2016ntu}
		ST-LSTM \cite{liu2016spatio}  & 69.2 & 77.7 & 1   \\ %\cite{liu2016spatio}
		STA-LSTM \cite{song2017end}  & 73.4 & 81.2 & 1  \\ %\cite{song2017end}
		VA-LSTM \cite{zhang2017view}  & 79.2 & 87.7 & 1   \\ %\cite{zhang2017view}
		ARRN-LSTM \cite{li2018skeleton}  & 80.7 & 88.8 & 2   \\ %\cite{li2018skeleton}
		\hline
		2s-3DCNN \cite{liu2017two}  & 66.8 & 72.6 & 2   \\ %\cite{liu2017two}
		TCN \cite{kim2017interpretable}  & 74.3 & 83.1 & 1   \\ %\cite{kim2017interpretable}
		Clips+CNN+MTLN \cite{ke2017new}  & 79.6 & 84.8 & 1   \\ %\cite{ke2017new}
		Synthesized CNN \cite{liu2017enhanced}  & 80.0 & 87.2 & 1   \\ %\cite{liu2017enhanced}
		3scale ResNet152 \cite{li2017skeleton} & 85.0 & 92.3 & 1  \\ % \cite{li2017skeleton}
		CNN+Motion+Trans \cite{li2017skeletonCNN}  & 83.2 & 89.3 & 2  \\ %\cite{li2017skeletonCNN}
        \hline
		ST-GCN \cite{yan2018spatial} & 81.5 & 88.3 & 1 \\ % (\cite{yan2018spatial})
		DPRL+GCNN \cite{tang2018deep} & 83.5 & 89.8 & 1 \\ % (\cite{tang2018deep}) 
		TA-GCN \cite{negarTAGCN} & 87.97 & 94.2 &  1 \\ % (\cite{negarTAGCN})
		AS-GCN \cite{li2019actional} & 86.8 & 94.2 & 2 \\ % (\cite{li2019actional})
		2s-AGCN \cite{shi2019two} & 88.5 & 95.1 & 2 \\ % (\cite{shi2019two})
		2s-TA-GCN \cite{negarTAGCN} & 88.5 & 95.1 & 2 \\ % (\cite{negarTAGCN})
		GCN-NAS \cite{peng2020learning} & 89.4 & 95.7 & 2 \\ % (\cite{peng2020learning})
		\hline \hline
		\bf{ST-BLN} & 85.71 & 93.80 & 1 \\ %
		\bf{2s-ST-BLN} & 87.8 & 95.1  & 2 \\ %
		\hline
    \end{tabular}}
\end{center}
\end{table}

\begin{figure}[!t]
    \centering
    \includegraphics[width=\linewidth]{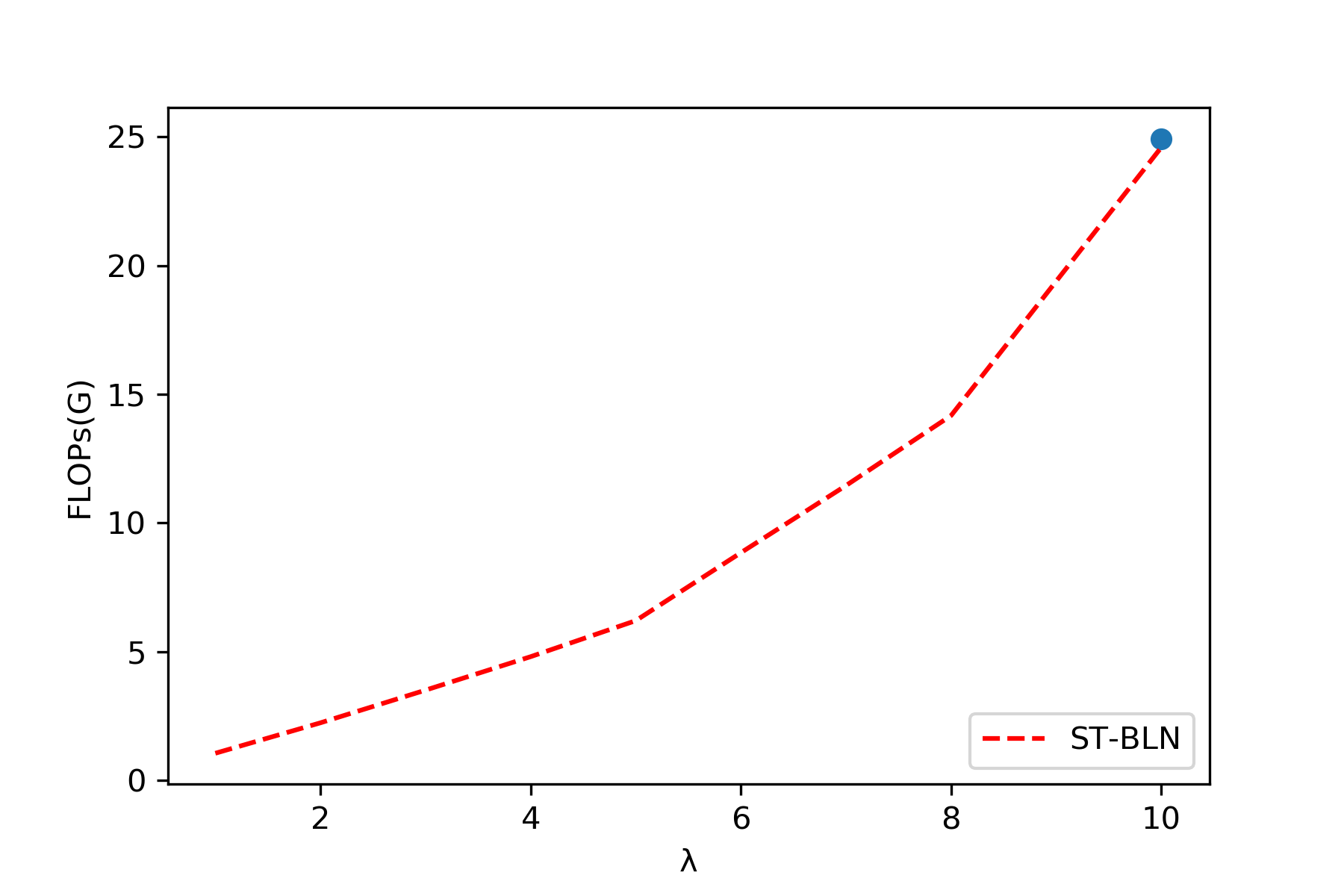}
    \caption{Floating Point Operations of the ST-BLN model when joints' features are aggregated in different layers $\lambda$ of the network. The case of a ST-BLN with $V^{(l)} = V, \:l=1,\dots,10$ is illustrated with a dot.} 
    \label{fig:FLOPs_BLN}
\end{figure}

To evaluate the need of combining information in the dimension of human body skeleton joints, expressed either as spatial attention of the form in Eq.  (\ref{eq:A2dConv_s}) and as bilinear mapping of the form in Eq. (\ref{eq:bl_Convs}), or if it is adequate the employ linear mappings effectively leading to neural layers combining a Multilayer Perceptron block with a Temporal Convolution block, we conducted a second set of experiments. In this set of experiments, we used a 10-layer spatio-temporal bilinear network formed by the same data transformation sizes as in the first set of our experiments, but instead of using bilinear mappings with $V^{(l)} = V$ for all 10 layers of the model, we use $V^{(l)} = V,\: l=1,\dots,\lambda -1$ and $V^{(l)} = 1,\: l=\lambda,\dots,10$. That is, the first $\lambda-1$ layers of the model use Spatio-Temporal Bilinear layers as the one illustrated in Fig. \ref{fig:ST-BLN1} with values $V^{(l)} = V$. At layer $\lambda$ a Spatio-Temporal Bilinear layer in the form of Fig. \ref{fig:ST-BLN1} is used with $V^{(\lambda)} = 1$, while the remaining layers take the form of a Spatio-Temporal Bilinear layer as the one illustrated in Fig. \ref{fig:ST-BLN2}. We applied experiments using both the CV and CS benchmarks of the NTU-RGB+D dataset using the joints data, and the results in terms of classification accuracy and number of FLOPs are indicated in Fig. \ref{fig:ACC_BLN} and Fig. \ref{fig:FLOPs_BLN}, respectively. As can be seen in Fig. \ref{fig:ACC_BLN}, on the CV benchmark the best performance is equal to $93.06\%$ and is obtained for a value of $\lambda = 6$. On the CS benchmark, the best performance is obtained for a value of $\lambda = 7$ and is equal to $84.47\%$. These performance values are on par with the results reported in Table \ref{tab:ACC_3variants}, achieved by using the Spatio-Temporal Bilinear/GCN layers using $V^{(l)} = V, \:l=1,\dots,10$. Observing the performance values achieved by the different networks in Fig. \ref{fig:ACC_BLN}, we can conclude that combining information in the dimension of human body skeleton joints at early layers of the network contributes to the performance achieved by the network, while this is not the case for deeper layers in the network. 

By observing the number of Floating Point Operations needed to evaluate an input skeleton sequence for different choices of $\lambda$ in Fig. \ref{fig:FLOPs_BLN}, we can see that one can increase efficiency of the model by using lower values of $\lambda$. This is due to that for layers beyond the $\lambda$-th layer of the network a smaller number of parameters is needed, as indicated by comparing the two layers illustrated in Fig. \ref{fig:ST-BLN1} and Fig. \ref{fig:ST-BLN2}. Considering that the number of model's parameters at the deeper layers is higher compared to the first layers, it can be seen that a considerable improvement in the model's efficiency can be achieved. Overall, a network using a value of $\lambda = 6$ ($\lambda = 7$) operates $\times2.78$ ($\times2.14$) faster compared to a network with $\lambda = 10$. As expected, the numbers of FLOPs for a network with $\lambda = 10$ and a ST-BLN with all layers using values $V^{(l)} = V$ are very similar, i.e. $24.59G$ and $24.93G$, respectively.

\section{Conclusion}\label{sec:conclusion}
In this paper, we studied the properties of the spatial attention used in ST-GCN methods for skeleton-based human action recognition. We made two observations, the first being that the spatial attention used in these methods leads to an asymmetric attention matrix. Our second observation is that the human-expert designed normalized Adjacency matrices used in models with additive attention serve only as an initialization process, and do not really affect parameters of the network for combining information in the dimension of human body skeleton joints used for inference. Based on the first observation, we proposed a symmetric spatial attention that can be directly incorporated in the existing ST-GCN methods without affecting the overall number of computations during inference. Based on the second observation, we proposed the Spatio-Temporal Bilinear Network (ST-BLN) which does not require the use of a predefined Adjacency matrix and allows for more flexible design of the model for skeleton-based human action recognition. Extensive experimental analysis shows that the three models lead to effectively the same performance. Moreover, by exploiting the flexibility provided by the proposed ST-BLN, one can increase the model's efficiency by a factor of more than $\times 2$ while preserving the performance levels of the original method. 

%\hl{should we cite the methods in table 3? there is no enough space for that}
%\hl{we have 1.5 pages extra}
%\hl{I should read the whole paper again, fill the first page, write the highlights, check references and check the author guide again}

\section*{Acknowledgment}
This work was supported by the European Union’s Horizon 2020 Research and Innovation Action Program under Grant 871449 (OpenDR). This publication reflects the authors’ views only. The European Commission is not responsible for any use that may be made of the information it contains.

\bibliographystyle{IEEEbib}
\bibliography{bibliography}

\end{document}